\documentclass[10pt,aps,prb,twocolumn,nofootinbib,longbibliography]{revtex4-1}
\usepackage{amssymb,amsmath}
\usepackage{subfig}
\usepackage{graphicx}
\usepackage{color}
\usepackage{hyperref}
\usepackage{listings}

\usepackage[ruled,vlined]{algorithm2e}

\hypersetup{
    bookmarks=true,         
    unicode=false,          
    pdftoolbar=true,        
    pdfmenubar=true,        
    pdffitwindow=false,     
    pdfstartview={FitH},    
    pdfauthor={Willie Hui Huang},     
    colorlinks=true,       
    linkcolor=blue,          
    citecolor=red,        
}
\graphicspath{{figures/}}

\begin{document}

\definecolor{dkgreen}{rgb}{0,0.6,0}
\definecolor{gray}{rgb}{0.5,0.5,0.5}
\definecolor{mauve}{rgb}{0.58,0,0.82}

\lstset{frame=tb,
  	language=Matlab,
  	aboveskip=3mm,
  	belowskip=3mm,
  	showstringspaces=false,
  	columns=flexible,
  	basicstyle={\small\ttfamily},
  	numbers=none,
  	numberstyle=\tiny\color{gray},
 	keywordstyle=\color{blue},
	commentstyle=\color{dkgreen},
  	stringstyle=\color{mauve},
  	breaklines=true,
  	breakatwhitespace=true
  	tabsize=3
}

\title{Performance evaluation and application of computation based low-cost homogeneous machine learning model algorithm for image classification}
\author{W. H. Huang}
\affiliation{Amazon Web Services, Seattle, WA, United States}
\date{August 27, 2020}

\begin{abstract}
The image classification machine learning model was trained with the intention to predict the category of the input image.  While multiple state-of-the-art ensemble model methodologies are openly available, this paper evaluates the performance of a low-cost, simple algorithm that would integrate seamlessly into modern production-grade cloud-based applications.  The homogeneous models, trained with the full instead of subsets of data, contains varying hyper-parameters and neural layers from one another.  These models' inferences will be processed by the new algorithm, which is loosely based on conditional probability theories.  The final output will be evaluated.  

\end{abstract}

\maketitle

\section{\label{sec:intro}Introduction}
The application of neural network based machine learning models on modern industrial sectors expanded tremendously [\onlinecite{1}] in recent years.  Specifically, researchers and engineers have great interests in image classification.  Utilizing the cost effectiveness of cloud-computing, massive quantities of software embed machine learning models as a portion of the cloud infrastructure.  Often times, the training of the models would be facilitated using cloud-based interfaces.  A pre-trained model could be utilized as a foundation for feature extraction and would potentially save training time while improving the model’s performance[\onlinecite{3}].  Other hyperparameters such as mini-batch size, learning rate, epoch and layer freezing for transfer learning are crucial factors to consider when the model is undergoing fine-tuning.

Section~\ref{sec:theory} illustrate the theoretical concepts that drove the research work as well as corresponding probability equations. Section~\ref{sec:process} states the experimental setup along with the respective hyperparameters. Section~\ref{sec:result} outlines the results of the methodology. Section~\ref{sec:discussion} provides arguments for the results observed and the application of the proposal in a cloud-based infrastructure.

\subsection{\label{sec:introProblem}Problem Statement}
Model ensembles are very popular techniques to improve the overall inferences' performances[\onlinecite{4}][\onlinecite{5}].  While multiple methodologies are openly available, the option of \textbf{only changing the computational algorithm} to process the models' prediction and yield improved result would be exceptionally attractive in the pragmatic engineering perspective.  

\subsection{\label{sec:introLit}Related Works}
The application of model ensemble was proposed to improve accuracy to various branches of machine learning[\onlinecite{6}].  The computational model ensemble technique of averaging the models' predictions was a simple yet effective approach to resolve variances between different model's predictions[\onlinecite{7}].  This paper is scoped to evaluate a novel computational technique that is loosely based on the conditional probability theorem for improving model ensemble performance.

\section{\label{sec:theory}Theoretical Framework}
The proposition to rank the models’ output predictions loosely based on the conditional probability theorem is the core concept that is applied to the post-processing of ensemble models in this work.  Given a test image for inference, a model would output a list of potential categories with the respective confidences.  As oppose to averaging the confidences for ensemble models and yield the category with the highest confidence, the proposition would be finding the category with lowest probability that the ensemble models inferred the image does not belong to the said category.  This section will walk-through the conceptual idea and the mathematical model behind the proposition.

Starting from the ensemble models, assume there are N trained models with different hyperparameters, ranging from number of layers to mini batch sizes, for N is a real integer and $>$ 2.  Given a list of predictions, P\textsubscript{1}, was yielded from model M\textsubscript{1} with validation accuracy of A\textsubscript{1}.  Another prediction list P\textsubscript{2} was yielded from model M\textsubscript{2} with validation accuracy of A\textsubscript{2}.  Finally, P\textsubscript{3} was yielded from model M\textsubscript{3} with validation accuracy of A\textsubscript{3}.  The first step of the proposition would be iterating through each class, C\textsubscript{j}, within P\textsubscript{1}, P\textsubscript{2}, P\textsubscript{3} and calculate the probability that the ensemble models inferred the test image not being class C\textsubscript{j}.  In another word, instead of gathering the models and finding out how confident the test image belongs to one class, the proposed method is finding how confident the test image does not belong to the said class.   

\begin{figure}[h]
\includegraphics[width=0.45\textwidth]{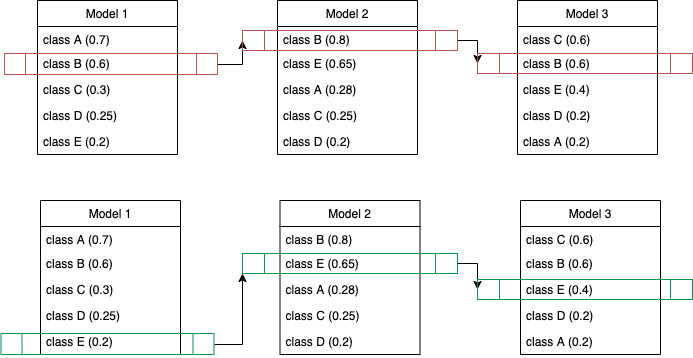}
\caption{The confidence values are in bracket for the different categories or classes}
\label{fig:ModifiedNMS}
\centering
\end{figure}
As referred in Figure~\ref{fig:ModifiedNMS}, each class in the prediction list is computed with the same class from other lists as shown with red and green boxes.  The confidence level is expressed in the form of probabilities with a maximum and minimum values being one and zero, respectively.\\
Before going further with the computation, it is assumed the models are trained with varying hyperparameters or neural architectures that allow larger variations in the neural network weights and biases between one another.  This is also an important criterion for ensemble models.  
Using Figure~\ref{fig:ModifiedNMS} example, the red rectangles of class B has three probabilities from three models.  In order to calculate the probability of the models correctly predict that the image is not class B, the following mathematical model is employed.

\begin{equation}
\label{eq:probOne}
P(C_1\cap T_1) = P(C_1 | T_1) P(T_1)
\end{equation}
Where T\textsubscript{1} refers to model 1, M\textsubscript{1}, is correct to predict a class and P(T\textsubscript{1}) equals A\textsubscript{1}
The conditional probability that M\textsubscript{1} predicted class C\textsubscript{1} is calculated via the probability of C\textsubscript{1} given T\textsubscript{1} multiply by the probability of T\textsubscript{1}
\begin{equation}
\label{eq:probTwo}
P(!C_{1_{e*}}) = 1 - \prod_{n=1}^N P((C_{1_{n}})\cap T_n) 
\end{equation}
Where e* means ensemble.
In theory, the probability of the ensemble correctly predict that the test image not being class C\textsubscript{1} would be calculated using Equation~\ref{eq:probTwo}.  However, the proposition is loosely based on the probability model, thus, some modifications were performed to the equations and yield the algorithm as shown in the form of pseudo-code in Appendix~\ref{sec:Aone}.

Notice that the maximum of the ensemble probabilities is picked for the class C\textsubscript{1} in the algorithm.  The concept is that the ensemble should estimate how unlikely the test image belongs to class C\textsubscript{1} and select the maximum probability.  Then the value will be compared with that of other categories.  In the end, the category with the minimal probability of all the categories will be selected as output for the test image classification.  Putting the concept into simple terms, the process could be thought of as a double negation.  First, find how unlikely the test image belongs to each individual category.  The lower the value would meant the more likely that the test image belongs to this category and vice versa.  Then, amongst all the computed outputs, finding the most probable category meant the minimal outputs would be retrieved.    

The quantitative values of the probabilities can only be used in the ranking and should not be referenced as the actual model confidence of the category.  Further limitations will be discussed in later sections of the paper. 

\section{\label{sec:process}Experimental Procedure}
Amazon Web Services (AWS) is employed for the experimentation and development of the work discussed in this paper.  Similar features or technologies could be interchanged or substituted with open source software or libraries.  

\subsection{\label{sec:dataLabel}Data Labelling}
Caltech256 and IMFDB images were used in the experimentation.  The data sets were segregated into three segments: training, validation and test.  As the images are already categorized, segmentation of images data set occurred from each category for training, validation and testing, respectively.  AWS S3 was used for storing the images into their respective folder categories and a python script was executed to pull the images into AWS SageMaker for training, validation and finally testing. 

\subsection{\label{sec:training}Model Training}
AWS Sagemaker was employed to facilitate the training of the image classification model using the image data set in AWS S3.  Similar training procedures can be performed using machine learning frameworks and libraries such as Tensorflow, PyTorch, Keras, Caffe, etc.  
Numerous runs were triggered using different hyper-parameters and six models were selected for the ensemble.  The models were trained with the same data-set.  The respective hyper-parameters for the models are listed in Table~\ref{tab:hyperparameters}.  All models use pre-trained model as basis.

\begin{table}[h]
\caption{\label{tab:hyperparameters}Hyperparameters for the ensemble models.}
\scriptsize
\begin{ruledtabular}
\begin{tabular}{*7l}
Model & One & Two & Three & Four & Five & Six\\ \hline
layers & 18 & 34 & 50 & 101 & 152 & 34 \\ \hline
weighted loss & true & false & false & true & true & true\\ \hline 
learning rate & 0.002 & 0.004 & 0.001 & 0.001 & 0.001 & 0.001 \\ \hline
mini batch size & 128 & 128 & 64 & 64 & 64 & 32 \\
\end{tabular}
\end{ruledtabular}
\end{table}

\subsection{\label{sec:deploy}Batch Transform and Post-processing}
The test data set were inferred by the six models using batch transform process.  The outputs were yielded in the format of lists with the list index corresponds to the category and the value being the confidence with a maximum and minimum value of one and zero, respectively.   
The outputs, or predictions, would be consumed by the proposed algorithm and yield the computed category.  The predicted category would be compared with the expected category for that test image and there would either be a match or a mismatch.  The result will be tallied up and divided by the total test data set and yield the ensemble performance as shown in Results~\ref{sec:result} Table~\ref{tab:results}.  The conventional method of averaging the models' predictions and yield the category with highest average confidence is also processed as a control comparison.

\section{\label{sec:result}Results}
The results of batch transforms from Section~\ref{sec:deploy} were processed and tallied to yield the overall performances of the conventional ensemble algorithm and the proposed algorithm.

\begin{table}[h]
\caption{\label{tab:results}Comparison of the ensemble algorithms and the respective performances}
\scriptsize
\begin{tabular}{ |p{5.7cm}|p{2.3cm}|  }
 \hline
 Methodology & Performance (\%) \\
 \hline
 \hline
 Top Performing Model.  & 73.56 \\
 \hline
 Averaging Prediction   & 77.88    \\
 \hline
 Proposed Algorithm   & 79.81    \\
 \hline
\end{tabular}    
\end{table}

Referring to Table~\ref{tab:results}, the proposed algorithm has an approximated 1\% increase in performance comparing with the conventional averaging algorithm based on the experimentation.

\section{\label{sec:discussion}Discussion}
Based on the result of the proposed algorithm for ensemble models, the boost in performance would be consider rather significant.  Only the computation algorithm is changed while the trained ensemble models were left untouched.  The advantages of the proposed technique in terms of operational perspective are as follows:
\begin{itemize}\parskip0pt
    	\item Does not require increasing the training data set
    	\medskip
    	\item Does not require further tuning of the hyper-parameters
\end{itemize} 
The conventional ensemble algorithm tends to simplify the variances of between the ensemble models' inferences, thus, overlooked subtle patterns within the output data.  The proposed algorithm provides more insight into the relationships between the predictions.  In addition, the validation performance of individual model was also taken into consideration as seen in Appendix~\ref{sec:Aone}.  This would act as a weighted coefficient to further fine tune the relationships of the models in the ensemble.  

Being able to use the models in their current state with only minor modifications to the overall system architecture would be a tremendous convenience in practice.  The proposed technique is a low cost alternative for improving the ensemble performance.  There are also obvious limitations when employing the proposed methodology and will be discussed as follows.

\subsection{\label{sec:limit}Limitation}
The theoretical model is loosely based on specific concepts of the probability theories.  Thus, the resulting probability values should only be used in a relative sense.  That is, the values serve as references for ranking the predicted categories.  Note that this same assumption holds true for the conventional ensemble algorithm.

\subsection{\label{sec:application}Application}
\begin{figure}[h]
\includegraphics[width=0.4\textwidth]{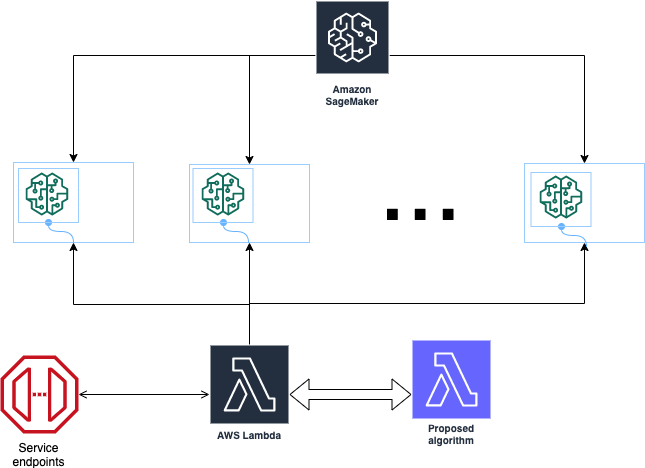}
\caption{Integrating the algorithm in Lambda}
\label{fig:CloudInf}
\centering
\end{figure}

Integration into cloud-based systems would be easily achieved via the usage of Lambda functions containing the proposed algorithm.  The advantage of the proposal became apparent since the technique is computational in nature and can be self-contained in a Lambda functions, which can be easily inserted into cloud infrastructure.  Referring to Figure~\ref{fig:CloudInf}
The models, trained by AWS SageMaker, are being encapsulated by running instances such that AWS Lambda function could invoke them and retrieve the inferences from an input image requested by the service endpoint.  The inferences would be processed by the proposed computational algorithm within another Lambda function and yield the final output.  The lambda would allow high concurrency and great scaling for the application.

\section{\label{sec:conclusion}Conclusion}
Image classification is a major field in the machine learning sector.  Other than tuning hyperparameters and increasing data set, ensemble models would be a great alternative to boost the overall performances.  This paper investigated a novel ensemble computational algorithm based loosely on probability theories and demonstrated a ~\% boost in performance comparing to the conventional computational ensemble technique.  Although there are limitations and boundaries with regards to the theoretical aspect of the proposed technique, it would be sufficient to apply to most applications in practice.

\appendix
\section{\label{sec:Aone}Proposed Algorithm in Pseudo-Code}
The following pesudo-code function would consume prediction lists from ensemble models and yield the predicted category.
\begin{lstlisting}[language={}]
/* In order to infer the category of the test image the arguments are the each model's list of confidences with the associated category.  For example, model_1_pred = [0.1, 0.2, 0.44, 0.33]  where the index of the list corresponds to the category which is ['class A', 'class B', 'class C', 'class D'], for this example there are a total of four classes. */

function ensembleProbabilityAlgorithm(model_1_pred, model_2_pred, ..., model_N_pred) {
    category_1_result
    category_2_result
    ...
    category_N_result

    for index in range(length of model_1_pred):
    /* iterate through each index (category) and compute the maximum probabilty that the models predict the test image not being the said category */    
        model_1_result = 1 - model_2_pred[index] * model_3_pred[index] * ... * model_N_pred[index] 
        + (1 - model_1_pred[index]) * model_1_accuracy
        /* model_1_result is equal to 1 -  (the probability of all the models correctly predict the category) plus the probability of model 1 predicting the test image not being the current category times the validation accuracy of model 1. */

        model_2_result = 1 - model_1_pred[index] * model_3_pred[index] * ... * model_N_pred[index] 
        + (1 - model_2_pred[index]) * model_2_accuracy
        /* notice that for model_2_result the third term is switched from model_1_pred[index] to model_2_pred[index] and the model 2's validation accuracy (model_2_accuracy). continue through all the models */
        
        ...

        model_N_result = 1 - model_1_pred[index] * model_2_pred[index] * ... * model_(N - 1)_pred[index] 
        + (1 - model_N_pred[index]) * model_N_acuracy


        category_(index)_result = the maximum of {model_1_result, model_2_result, ..., model_N_result}
        /* the maximum of the probability for the models prediction of the test image not being in this category is stored outside of the loop */

    return the index of the minimum of {category_1_result, category_2_result, ..., category_N_result}
    /* we are returning the category with the least probability value of the models "think" they predicted the test image not being this category this is a double negation. In another word, for each category, how confident are the models think the test image is NOT this category, then find the category with the smallest values (least confident that this is not the category).  The double negation will yield the category for the test image */
}
\end{lstlisting}

\bibliography{bib}
\end{document}